\pdfoutput=1

\documentclass[11pt]{article}

\usepackage[preprint]{acl}

\usepackage{times}
\usepackage{latexsym}

\usepackage[T1]{fontenc}

\usepackage[utf8]{inputenc}

\usepackage{microtype}

\usepackage{inconsolata}

\usepackage{graphicx}
\usepackage{float}

\usepackage{amsmath}
\usepackage{amssymb} 
\usepackage{amsfonts} 
\usepackage{enumitem}
\usepackage{bm} 
\usepackage{mathtools} 
\usepackage{booktabs}
\usepackage[final]{changes}

\usepackage{multicol}
\usepackage{subcaption}
\usepackage{booktabs}
\usepackage{multirow}
\usepackage{float}

\usepackage{etoolbox}
\AtBeginEnvironment{table}{\scriptsize}

\usepackage{tikz}
\usetikzlibrary{positioning, shapes.geometric, arrows.meta, fit, backgrounds}
\usetikzlibrary{fit}

\title{Inceptive Transformers: Enhancing Contextual Representations through Multi-Scale Feature Learning Across Domains and Languages}

\author{Asif Shahriar\textsuperscript{1,2}, Rifat Shahriyar\textsuperscript{1}, M Saifur Rahman\textsuperscript{1}\\
\textsuperscript{1}Bangladesh University of Engineering and Technology,  
\textsuperscript{2}BRAC University \\
\texttt{asif.shahriar@bracu.ac.bd, \{rifat, mrahman\}@cse.buet.ac.bd}}

\begin{document}
\maketitle
\begin{abstract}

Encoder transformer models compress information from all tokens in a sequence into a single \textbf{\texttt{[CLS]}} token to represent global context. This approach risks diluting fine-grained or hierarchical features, leading to information loss in downstream tasks where local patterns are important. To remedy this, we propose a lightweight architectural enhancement: an inception-style 1-D convolution module that sits on top of the transformer layer and augments token representations with multi-scale local features. This enriched feature space is then processed by a self-attention layer that dynamically weights tokens based on their task relevance. Experiments on five diverse tasks show that our framework consistently improves general-purpose, domain-specific, and multilingual models, outperforming baselines by 1\% to 14\% while maintaining efficiency. Ablation studies show that multi-scale convolution performs better than any single kernel and that the self-attention layer is critical for performance.

\end{abstract}

\section{Introduction}\label{sec:intro}

\begin{figure*}[!htb]
    \centering
    \begin{subfigure}[b]{0.42\textwidth}
        \centering
        \includegraphics[width=\textwidth]{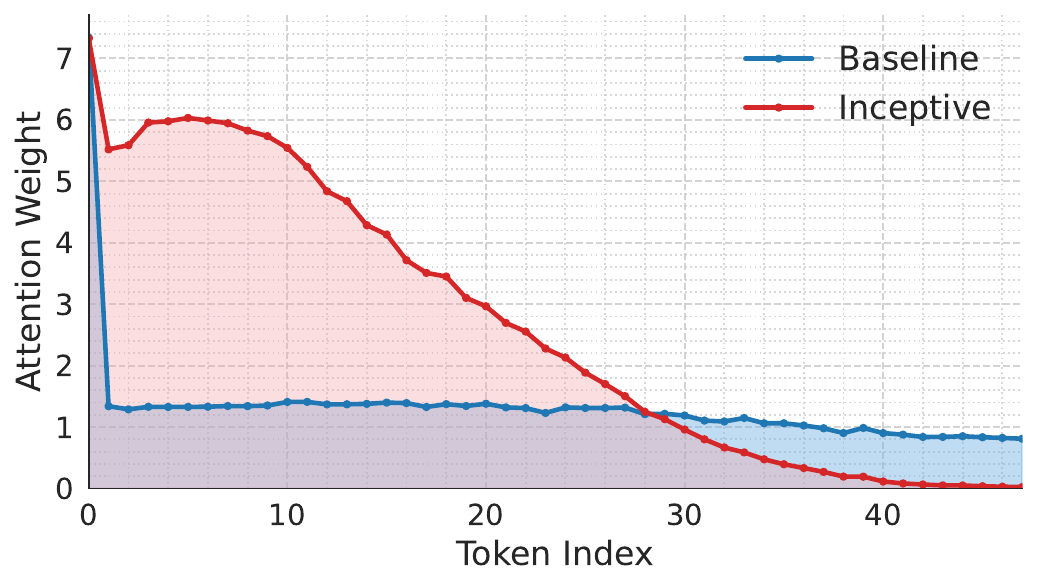}
        \caption{Baseline vs Inceptive DeBERTa v3 (Irony)}
        \label{subfig:intro-db}
    \end{subfigure}
    \hfill
    \begin{subfigure}[b]{0.28\textwidth}
        \centering
        \includegraphics[width=\textwidth]{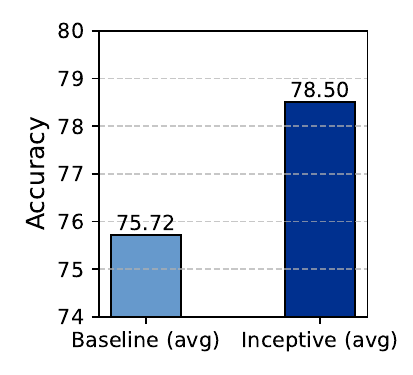}
        \caption{Irony Detection}
        \label{subfig:intro-irony}
    \end{subfigure}
    \hfill
    \begin{subfigure}[b]{0.28\textwidth}
        \centering
        \includegraphics[width=\textwidth]{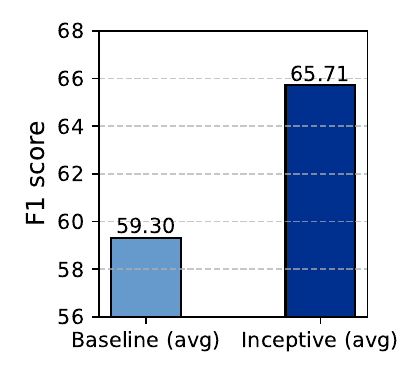}
        \caption{OHSUMED}
        \label{subfig:intro-ohsumed}
    \end{subfigure}

    \caption{Encoder models like DeBERTa over-rely on the [CLS] token, whereas inceptive DeBERTa redistributes attention dynamically based on task contribution (\ref{subfig:intro-db}). Experiments show the superior performance of inceptive models (\ref{subfig:intro-irony} and \ref{subfig:intro-ohsumed}).}
    \label{fig:intro}
\end{figure*}

Since its introduction, the transformer architecture~\cite{vaswani2017attention} has revolutionized the field of natural language processing (NLP). Encoder-based transformer models such as BERT~\cite{devlin2018bert}, RoBERTa~\cite{liu2019roberta}, XLNet~\cite{yang2019xlnet}, Electra~\cite{clark2020electra}, DeBERTa v3~\cite{he2021debertav3}, and ModernBERT~\cite{warner2024modernbert} have demonstrated impressive performance across a wide range of NLP tasks. In addition to these general-purpose models, a number of domain-specific BERT-based models like BioBERT~\cite{lee2020biobert}, SciBERT~\cite{beltagy2019scibert}, LegalBERT~\cite{chalkidis2020legalBERT}, BERTweet~\cite{nguyen2020bertweet} have emerged, which are further pre-trained on domain-specific corpora to capture the unique language, terminology, and stylistic features of various specialized fields. In parallel, cross-lingual models like XLM-R~\cite{conneau-2020-xlmr} and language-specific models such as BanglaBERT~\cite{bhattacharjee-2022-banglabert} have extended this architecture to support diverse linguistic settings, including low-resource languages like Bangla.

All these encoder models are designed to aggregate all token embeddings into a single representation, called the \textbf{\textbf{\texttt{[CLS]}}} token, which is later used for downstream tasks like classification. Although convenient, this approach of collapsing an entire text sequence with multiple aspects into one single embedding can cause information loss~\cite{chang2023multi}, particularly when there are short-range dependencies between tokens~\cite{Guo_Zhang_Liu:Gaussian_transformer:2019, act:2021}. During experiments, we observed that the over-reliance on \textbf{\texttt{[CLS]}} token makes the encoder models insufficient in capturing fine-grained contextual nuances or localized cues critical for tasks like emotion recognition or irony detection (Fig. \ref{subfig:intro-db}). This issue is even more pronounced in multi-label tasks, which require token-level attention rather than a single sequence-level summary.

To address these limitations, we propose \textit{Inceptive Transformers} -- a lightweight and modular architecture that augments a transformer baseline by stacking an inception-style 1-D convolution module on top. Instead of using \textbf{\texttt{[CLS]}}-based pooling, we feed the final hidden states from the baseline transformer (e.g. RoBERTa or BioBERT) to a multi-scale feature extraction module inspired by inception networks. This module employs parallel 1-D convolutional filters with varying kernel sizes that are designed to recognize local features, such as key phrases or word combinations that are indicative of specific classifications. The goal of the inception module is to incorporate local features without sacrificing global context, which is achieved by using a residual connection to concatenate the original transformer's hidden states with the multi-scale features. These enriched features are then processed by a self-attention mechanism, which dynamically assigns weights to tokens based on their task-specific contribution, thus allowing the model to effectively prioritize relevant tokens. 

As an illustration, let us consider the ironic text: \emph{``work Wednesday--Sunday\ldots\ \#yay \#not \#moneytho''}. BERTweet fails to identify it as ironic because the dominant neutral phrase (\emph{``work Wednesday--Sunday''}) outweighs the local sarcastic hashtags when collapsed into a single \textbf{\texttt{[CLS]}} representation. In contrast, Inceptive BERTweet correctly predicts irony because its multi-scale convolution and post-hoc attention redistribute focus toward the localized cues (\emph{`\#yay', `\#not'}), capturing the contradiction that signals sarcasm.

Our experiments demonstrate that Inceptive Transformers consistently outperform both general-purpose (RoBERTa, DeBERTa v3, ModernBERT, XLM-R) and domain-specific (BERTweet, BioBERT, CT-BERT, BanglaERT) baselines. On five different tasks (Bangla and English emotion recognition, irony detection, disease identification, and anti-vaccine concern classification), we observed performance gains from \textbf{1\%} to as high as \textbf{14\%}, with less than \textbf{10\%} inference-time overhead. Attention maps show that inceptive models redistribute attention from [CLS] toward task-relevant spans, increasing token coverage and mitigating over-reliance on any single token.

Major contributions of our work are as follows.
\begin{itemize}[leftmargin=*]
    \item \textbf{A Novel, Modular Architecture.} We introduce a lightweight, modular, plug-and-play architecture for enriching the contextual representations of transformer models through a post-hoc multi-scale feature extraction module that can be integrated into any pre-trained encoder model without costly re-training. To the best of our knowledge, this has not been explored before.
    \item \textbf{A Generalizable Framework.} We propose a domain- and backbone-agnostic framework that can be applied on top of any encoder, including general-purpose, domain-specific, and cross-lingual models, without any model-specific adjustments. Furthermore, through comprehensive evaluations, we show that our inceptive models perform strongly in four different datasets from diverse domains and languages -- highlighting its general applicability.
    \item \textbf{Rigorous Validation.} We validate the effectiveness of our method through a rigorous empirical evaluation that includes statistical significance testing across multiple runs and extensive ablation studies for isolating the impact of each architectural component.
\end{itemize}

\section{Related Work}\label{sec:related_works}

\textbf{Text classification models} range from traditional machine learning approaches like decision trees~\cite{Law:DT_multi_label:2022}, support vector machines (SVM), k-nearest neighbors (KNN)~\cite{HANIFELOU:kNN:2018}, and ensemble learning~\cite{zhu:dynamin_ensemble:2023, wu:ml_forest:2016}, to more advanced deep learning techniques like RNN and LSTM \cite{Lai_Xu_Liu_Zhao_2015, Onan2022BidirectionalCR}. Convolution networks have also been used \cite{conneau-etal-2017-deep, choi-etal-2019-adaptive, Yao_Mao_Luo_2019, Soni2022TextConvoNetAC}, but they often struggle to capture long-range dependencies in text.

\noindent \textbf{Combining convolution with transformers.} After the transformer architecture \cite{vaswani2017attention} was introduced, many works have combined convolution with transformers, but these works mostly focus on vision related tasks \cite{Fang_2022_CVPR, si2022inceptiontransformer, YUAN2023109228}. Application on NLP domain remains limited to a few works \cite{zheng:bert_cnn:2019, wan2022financial, chen:chinese_bert:2022, wu2024xlnetcnngre} — which mostly focus on improving a particular transformer model, like BERT or XLNet. In comparison, we provide a general architecture capable of improving different types of transformer models, both domain-specific and general-purpose.

\noindent \textbf{Modifications of BERT-like models.} Several works modify BERT through architectural or pretraining adaptations to better suit specific tasks or domains, including SpanBERT~\cite{joshi2020spanbert}, StructBERT~\cite{wang2019structbert}, and CodeBERT~\cite{feng2020codebert}. Other works such as MT-DNN~\cite{liu2019mtdnn} introduce multi-task learning objectives on top of BERT, while KnowBERT~\cite{peters2019knowbert} integrates external knowledge bases into BERT’s architecture. Our work is orthogonal to these efforts: instead of modifying the pre-training strategy, we propose an architectural enhancement that can be directly plugged into existing BERT-like models.

\noindent \textbf{Encoder vs LLMs.} While LLMs have shown impressive generative capabilities in recent years, a number of studies demonstrate that encoder models still perform better on supervised non-generative tasks like classification and NER. ~\citet{edwards:incontext:2024} compared in-context prompting with LLaMA/Flan-T5 against fine-tuning encoder models like RoBERTa on 16 datasets and found that fine-tuned encoder models performed better. \citet{sun:carp:2023} showed that even with few-shot prompting and chain-of-thought reasoning, LLMs underperformed compared to fine-tuned models like RoBERTa and XLNet in classification tasks. \citet{chen2025benchmarking} found that encoder models like BioBERT and PubMedBERT remain state-of-the-art in biomedical domain tasks like document classification, NER, and relation extraction –- significantly outperforming LLMs like GPT-4 and LLaMA 2 13B.


    

    

    \begin{figure*}[t]
        \centering
        \includegraphics[width=\textwidth]{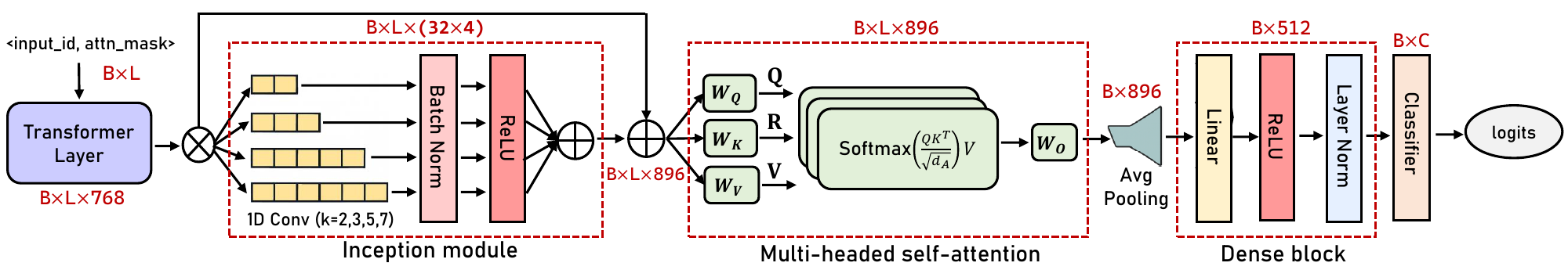}
        \caption{End-to-end architecture of the Inceptive Transformer framework. Output shapes are annotated for each main component. $B: $ batch size, $L:$ input sequence length, $C:$ number of classes. In this figure we have used 768 as hidden state dimension of BERT-like transformer models, 32 as output channels for each convolution branch, and 512 as the target dimensionality of the linear layer in dense block. $W_Q, W_K, W_V, W_O$ are learnable projection matrices, and $d_A$ is the dimension of each attention head.}
        \label{fig:model-architecture}
    \end{figure*}

\section{Inceptive Transformer}\label{sec:description}


\subsection{Motivation}\label{subsec:motivation}

Transformer-based models rely on token-level embeddings derived primarily from self-attention layers to capture global dependencies and context within text sequences. In our experiment, we visualized the attention maps of these models in Section \ref{subsec:analysis}, which show a strong bias in attention towards the \textbf{\texttt{[CLS]}} token, while intermediate tokens often receive comparatively lower attention.  The \textbf{\texttt{[CLS]}} token is a weighted aggregation of all token embeddings in the sequence, which the model relies on to represent the entire sequence for classification tasks. This bias suggests an underutilization of contextual and local dependencies, potentially limiting the model’s ability to effectively capture fine-grained patterns and hierarchical structures present in textual data.


Our model is designed to address this gap by incorporating convolutional operations, which excel at capturing local patterns and hierarchical structures in data~\cite{gu2018cnnadvancements, Li2022CNNsurvey}. CNNs are typically not used on textual data due to their inability to capture long-range dependencies. However, using convolution makes sense in our model because it operates on embeddings generated by a transformer— not on raw text. This allows the convolutional operations to refine the already globally contextualized embeddings by emphasizing fine-grained, local features that might otherwise be overlooked. Furthermore, instead of using a single convolution layer with a fixed kernel size, we use an inception module~\cite{Szegedy2015inception} to apply convolutions with multiple kernel sizes to learn features at different levels of granularity-- capturing both token-level patterns and phrase-level dependencies.


The applicability of our model is not limited to general-purpose transformers like RoBERTa. Domain-specific pre-trained models such as BioBERT, CT-BERT, or BERTweet show similar attention biases as BERT and RoBERTa, leading to challenges in capturing local and hierarchical dependencies. By integrating our model’s multi-scale feature extraction approach, these domain-specific variants can also be enhanced, improving their ability to represent diverse patterns within specialized input data. This versatility makes our model a robust addition to any transformer-based architecture. 



\subsection{Model Architecture}\label{subsec:architecture}

The full workflow of our inceptive models is illustrated in Fig.\ref{fig:model-architecture}. The input to our model is preprocessed text data, which need to be tokenized using an appropriate pre-trained tokenizer corresponding to the chosen transformer model. Mathematically, the input can be represented as $X = [x_1, x_2, \dots, x_L]$ where \( L \) is the sequence length, and each \( x_i \) corresponds to a token from the text. $X$ is passed to the transformer layer.

\subsubsection{Transformer Layer}

The first layer of our architecture is a transformer-based model, such as RoBERTa or BioBERT. \added{Given input \( X \), the transformer layer generates a tensor of hidden states $H \in \mathbb{R}^{B \times L \times d}$ where \( B \) is the batch size, \( L \) is the sequence length, and \( d \) is the hidden state dimension. We denote \( H[b, i, :] = h_i^{(b)} \in \mathbb{R}^d \) as the contextual embedding for the \( i \)-th token in the \( b \)-th input.} A dropout layer is applied to \( H \) to prevent overfitting.

\subsubsection{Inception Module}

The primary task of this layer is to extract multi-scale local features. The inception module receives contextual embeddings $H$ generated by the transformer and applies parallel convolutional layers with small kernel sizes $k$ (e.g., $k = 2, 3, 5, 7$) to learn features at different granularities.  Smaller kernels ($k = 2 \text{ or } 3$) capture fine-grained token-level relationships, such as modifiers or word pair dependencies, whereas larger kernels ($k = 5 \text{ or } 7$) capture slightly broader local patterns, such as syntactic or semantic relationships over small phrases.


\added{Each branch of the inception module applies a 1D convolution over the sequence of contextual embeddings generated by the transformer. Let the input be \( H \in \mathbb{R}^{L \times d} \), where \( L \) is the sequence length and \( d \) is the hidden size. For a convolution with kernel size \( k \), each filter has weights \( W \in \mathbb{R}^{k \times d} \) and a bias term \( b \in \mathbb{R} \). The output at position \( i \) is computed as:
\begin{equation*}
    Y_i = \sum_{j=0}^{k-1} W_j \cdot H_{i + j} + b
\end{equation*}
where \( H_{i + j} \in \mathbb{R}^d \) is the embedding of the \( (i + j) \)-th token, and \( W_j \in \mathbb{R}^d \) is the \( j \)-th row of the filter. This operation slides across the sequence to produce a feature map \( Y \in \mathbb{R}^{L \times c} \), where \( c \) is the number of convolutional filters (i.e. output channels) used in the branch. To preserve the original sequence length, we apply appropriate padding: for kernel size 2, we use right padding of 1; for kernel sizes 3, 5, and 7, we apply symmetric padding.}

After the convolution, each branch further processes its output using batch normalization to stabilize and accelerate the training process, followed by a ReLU activation to introduce non-linearity. Finally, the outputs of all four branches are concatenated along the channel dimension to form a combined feature map $C \in \mathbb{R}^{B \times L \times (4 \cdot c)}$. To preserve information from the original transformer output, we concatenate \( H \) and \( C \) along the feature dimension to form $R \in \mathbb{R}^{B \times L \times (d + 4c)}$. This residual connection ensures that the original features are retained alongside the multi-scale features. This combined representation, enriched with both global and multi-scale local features, is then passed to the self-attention layer for further processing.

\subsubsection{Self-Attention}

After the inception module extracts multi-scale features, an additional self-attention mechanism is necessary to capture dependencies and relationships across the enriched feature space $R$. This ensures that tokens that contribute the most to the task are effectively prioritized and selected, allowing the model to focus on the most relevant features.



Given $R \in \mathbb{R}^{B \times L \times d_R}$, the attention mechanism maps it to query $Q$, key $K$, and value $V$:
\[
Q = R W_Q, \quad K = R W_K, \quad V = R W_V
\]
where $W_Q, W_K, W_V \in \mathbb{R}^{d_R \times d_A}$, $d_R$ is the enriched feature space dimension, and $d_A$ is the attention head dimension. The attention scores are computed as:
\[
\text{Attention}(Q, K, V) = \text{softmax}\left(\frac{QK^\top}{\sqrt{d_A}}\right)V
\]
Since we use multi-headed attention, the outputs of multiple attention heads are concatenated and projected back to the original feature space:
\[
A = \text{Concat}(\text{head}_1, \ldots, \text{head}_h) W_O
\]
where $W_O \in \mathbb{R}^{(h \cdot d_A) \times d_R}$ is a learnable projection matrix and $h$ is the number of attention heads, another tunable hyperparameter. The attention output $A \in \mathbb{R}^{B \times L \times d_R}$ captures refined dependencies across both token positions and feature scales.



\subsubsection{Adaptive Average Pooling}

To reduce the sequence-level representation \(A\) to a fixed-size vector suitable for classification, global average pooling is applied across the sequence length. Given the attention output \(A \in \mathbb{R}^{B \times L \times d_R}\), we first permute it to \(\mathbb{R}^{B \times d_R \times L}\). Afterwards, adaptive average pooling computes the average over the entire sequence for each feature channel, regardless of the input length, by dynamically adjusting the pooling region. Mathematically:
\[
P_{b,i} = \frac{1}{L} \sum_{j=1}^{L} a_{b,i,j}
\]
where \(a_{b,i,j}\) is the value of the \(i\)th feature channel at the \(j\)th position for the \(b\)th example. This produces a tensor \(P \in \mathbb{R}^{B \times d_R \times 1}\), which is then squeezed to yield a final pooled representation \(P \in \mathbb{R}^{B \times d_R}\).

\subsubsection{Dense Block}

For further refinement, the pooled representation $P$ is passed through a dense block consisting of three sublayers. First, a fully connected layer is used to reduce the dimensionality by $D = P W_d + b_d$ where $W_d \in \mathbb{R}^{d_R \times d_D}$, $b_d \in \mathbb{R}^{d_D}$, and $d_D$ is the target dimensionality (e.g., 512). ReLU activation is used to introduce non-linearity, and layer normalization is used to stabilize the output. The output of the dense block $D \in \mathbb{R}^{B \times d_D}$  represents a compact and refined feature set ready for classification.

\subsubsection{Final Classification}

The output $D$ is passed to a linear classifier, which computes logits for each class as $O = D W_f + b_f$; where $W_f \in \mathbb{R}^{d_D \times C}$ and $b_f \in \mathbb{R}^{C}$. The logits $O \in \mathbb{R}^{B \times C}$ are interpreted based on the task. 

\noindent\textbf{Notation.} For brevity, we will refer to our inceptive models using the convention $i-\text{Baseline}-n$, where $i$ signifies `Inceptive', `Baseline' refers to the Pretrained Model (PLM) (e.g. RoBERTa, BioBERT), and $n$ denotes the number of output channels in each branch of the inception module. For instance, \textit{i}BioBERT-128 is the Inceptive BioBERT model with 128 output channels per convolution.


\section{Experimental Setup}\label{sec:methodology}

\subsection{Datasets}

For a robust evaluation, we test our framework on five tasks from four datasets spanning diverse domains, text lengths, and both multi-class and multi-label settings. Multi-class tasks include emotion recognition~\cite{mohammad2018emotion} and irony detection~\cite{van2018irony} from the TweetEval benchmark~\cite{barbieri2020tweeteval}, alongside a large-scale Bengali emotion detection dataset~\cite{faisal:bangla:2024} to evaluate performance on a morphologically rich, low-resource language. Multi-label tasks include OHSUMED \footnote{\href{https://disi.unitn.it/moschitti/corpora.htm}{OHSUMED-link}}, a collection of biomedical journal abstracts, and CAVES \cite{poddar2022caves}, a dataset of anti-COVID vaccine concerns such as side-effects, ingredients, corporate greed, political motivations, etc. More details on the datasets can be found in Appendix \ref{appendix:datasets}.

\subsection{Model Training and Evaluation}

Each input sequence is tokenized using a model-specific tokenizer and then passed through the model to generate logits. For multi-class classification, the model predicts mutually exclusive class probabilities using softmax activation and cross-entropy loss, whereas for binary and multi-label tasks, it outputs non-exclusive probabilities with sigmoid activation and binary cross entropy with logits loss. During backpropagation, gradients were clipped to a maximum norm of 1.0 to ensure numerical stability. The AdamW optimizer~\cite{kingma2015adam} with weight decay was used to update the model weights.

The training process was conducted iteratively over multiple epochs, with a Cosine Annealing learning rate scheduler. At the end of each epoch, the model was evaluated on the validation dataset to monitor key metrics, including accuracy, F1-score, AUC-ROC (multi-class), AUPR (multi-label), and inference time. The best model was selected based on accuracy for binary and multi-class classification tasks, and F1-score for multi-label tasks. Each model was run 10 times on each dataset. The models were trained and evaluated using 40GB A100 GPU. However, all of our models can be run on 16 GB GPUs (e.g. P100). We used the \texttt{transformer} version 4.48.3.

\subsection{Hyperparameters}

\begin{table}[h!]
\centering
\caption{Hyperparameters.}
\begin{tabular}{lc}
\toprule
Hyperparameter & Value \\
\midrule
Sequence Length & 128, 512 (ohsumed) \\
Batch Size & 32 \\
Epochs & 12 \\
Learning Rate & 1e-5 \\
Weight Decay & 1e-3, 1e-4 (ohsumed, caves) \\
Sigmoid threshold & 0.5 \\
\bottomrule
\end{tabular}
\label{table:hyperparameters}
\end{table}

The hyperparameters used in this experiment are shown in Table \ref{table:hyperparameters}. \added{All hyperparameter values were selected empirically based on validation set performance.}

\section{Results}\label{sec:results}

\subsection{Comparative Performance}\label{subsec:comparison}

In this section, we compare the performance of the inception-enhanced models against the baselines. Multi-class performance comparison (in terms of accuracy) is shown in Table \ref{tab:multi-class}, while multi-label comparison (F1-score) is shown in Table \ref{tab:multi-label}. A detailed comparison can be found in appendix \ref{sec:appendix:full-perf}, where we also report metrics like precision, recall, AUC-ROC and AUPR, that also account for class imbalance. To account for stochasticity, we performed 10 independent training and evaluation runs for each model on each dataset, and reported the average performance on the test set. Performance comparison across all runs can be found in Appendix \ref{sec:appendix:all-runs}.

\begin{table}[!h]
    \centering
    \renewcommand{\arraystretch}{1.2}
    \setlength{\tabcolsep}{10pt}
    \caption{Multi-class performance comparison in test set}
    \small
    \begin{tabular}{lcp{1.6cm}}
        \toprule
        \textbf{Model} & \textbf{Accuracy} & \textbf{Inference Time (s)} \\
        \midrule
        \midrule
        \multicolumn{3}{c}{\textbf{Irony Detection}} \\
        \midrule
        BERTweet & 82.69 & 1.59 \\
        \textbf{\textit{i}}BERTweet-16 & \textbf{84.51} & 1.62 \\
        RoBERTa & 75.15 & 1.60 \\
        \textbf{\textit{i}}RoBERTa-32 & \textbf{77.08} & 1.68 \\
        ModernBERT & 67.77 & 1.85 \\
        \textbf{\textit{i}}MB-32 & \textbf{70.41} & 1.93 \\
        DeBERTa v3 & 77.27 & 1.89 \\
        \textbf{\textit{i}}DB v3-16 & \textbf{82.02} & 1.96 \\
        \midrule
        \midrule
        \multicolumn{3}{c}{\textbf{Emotion Recognition}} \\
        \midrule
        BERTweet & 83.29 & 2.83 \\
        \textbf{\textit{i}}BERTweet-64 & \textbf{84.11} & 2.93 \\
        RoBERTa & 81.69 & 2.88 \\
        \textbf{\textit{i}}RoBERTa-16 & \textbf{82.42} & 3.00 \\
        ModernBERT & 76.10 & 3.41 \\
        \textbf{\textit{i}}MB-16 & \textbf{78.70} & 3.47 \\
        DeBERTa v3 & 83.93 & 3.40 \\
        \textbf{\textit{i}}DB v3-16 & \textbf{84.16} & 3.55 \\
        \midrule
        \midrule
        \multicolumn{3}{c}{\added{\textbf{Bangla Emotion Recognition}}} \\
        \midrule
        BanglaBERT & 69.98 & 15.65 \\
        \textbf{\textit{i}}BanglaBERT-16 & \textbf{70.74} & 16.62 \\
        XLM-RoBERTa & 65.91 & 15.42 \\
        \textbf{\textit{i}}XLMR-16 & \textbf{66.53} & 15.77 \\        
        \bottomrule
    \end{tabular}
    \label{tab:multi-class}
\end{table}

\begin{table}[!h]
    \centering
    \renewcommand{\arraystretch}{1.2}
    \caption{Multi-label performance comparison in test set}
    \small
    \begin{tabular}{lcp{1.6cm}}
        \toprule
        \textbf{Model} & \textbf{F1-score} & \textbf{Inference Time (s)} \\
        \midrule
        \midrule
        \multicolumn{3}{c}{\textbf{OHSUMED}} \\
        \midrule
        BioBERT & 63.50 & 53.88 \\
        \textbf{\textit{i}}BioBERT-128 & \textbf{72.34} & 58.74 \\
        BioBERT-Large & 73.12 & 154.00 \\
        \textbf{\textit{i}}BioBERT-Large-128 & \textbf{74.77} & 158.01 \\
        RoBERTa & 61.53 & 67.42 \\
        \textbf{\textit{i}}RoBERTa-128 & \textbf{65.44} & 74.44 \\
        ModernBERT & 56.71 & 72.95 \\
        \textbf{\textit{i}}MB-128 & \textbf{64.07} & 77.17 \\
        DeBERTa v3 & 55.47 & 83.27 \\
        \textbf{\textit{i}}DB v3-128 & \textbf{60.98} & 87.57 \\
        \midrule
        \midrule
        \multicolumn{3}{c}{\textbf{CAVES}} \\
        \midrule
        CT-BERT & 74.24 & 10.27 \\
        \textbf{\textit{i}}CTBERT-16 & \textbf{74.86} & 10.56 \\
        BioBERT-Large & 71.00 & 10.75 \\
        \textbf{\textit{i}}BioBERT-Large-16 & \textbf{72.32} & 11.02 \\
        RoBERTa & 71.11 & 4.67 \\
        \textbf{\textit{i}}RoBERTa-32 & \textbf{72.11} & 4.78 \\
        ModernBERT & 63.31 & 5.01 \\
        \textbf{\textit{i}}MB-32 & \textbf{66.63} & 5.09 \\
        DeBERTa v3 & 69.05 & 5.05 \\
        \textbf{\textit{i}}DB v3-32 & \textbf{71.66} & 5.12 \\
        \bottomrule
    \end{tabular}
    \label{tab:multi-label}
\end{table}


In the task of detecting irony in social media posts, the domain-specific BERTweet model saw a \textbf{2.20\%} improvement through our inceptive framework, while the inference time increased 1.89\% -- highlighting a clear positive trade-off. Similarly, general-purpose encoder models like RoBERTa, DeBERTa and ModernBERT also improved by margins of \textbf{2.57\%}, \textbf{6.15\%}, and \textbf{3.90\%} respectively -- all of which were greater than the inference overhead incurred. Specifically, Inceptive DeBERTa performed at a level similar to that of BERTweet, despite the latter being domain-pretrained. 

In comparison, the improvements were modest in emotion recognition, where both Inceptive BERTweet-64 and Inceptive RoBERTa-16 achieved approximately \textbf{1\%} improvement over baselines. A more substantial gain was observed for ModernBERT, where the inceptive variant improved accuracy by \textbf{3.42\%}.  In Bangla emotion recognition, Inceptive BanglaBERT-16 improved on its baseline by \textbf{1.08\%}, while Inceptive XLM-RoBERTa-16 achieved a \textbf{0.94\%} increase in accuracy over vanilla XLM-RoBERTa .


On the long-form, multi-label OHSUMED dataset, our framework delivered its most significant improvements. Inceptive BioBERT achieved an average F1-score of 72.34, which is a \textbf{13.92\%} improvement over BioBERT that came at a cost of 9.02\% increase in runtime. Other models also benefited substantially: Inceptive ModernBERT showed a \textbf{12.98\%} F1-score increase, while Inceptive DeBERTa improved by \textbf{9.93\%}. Notably, Inceptive RoBERTa (65.44) performed better than BioBERT (63.50), which is specifically pre-trained on biomedical corpora; thus highlighting the generalization capability of our plug-and-play architecture. Furthermore, Inceptive BioBERT performed at a similar level as BioBERT-large, despite the latter taking almost 3x as much time to run and requiring significantly more compute power. Nonetheless, Inceptive BioBERT also improved by \textbf{2.26\%}, confirming the framework's applicability to larger models as well. 

Finally, our inceptive module demonstrated consistent gains in the noisy and complex CAVES dataset as well. 
BioBERT-large improved by \textbf{1.86\%} whereas CT-BERT-large saw a gain of \textbf{0.84\%}. The performance uplift was more pronounced among general-purpose models: RoBERTa improved by \textbf{1.41\%}, DeBERTa by \textbf{3.78\%}, and ModernBERT by \textbf{5.24\%}.

Overall, our lightweight inceptive framework has improved every PLM we tested while maintaining reasonable efficiency. The performance delta varies depending on the baseline capability. General-purpose models like DeBERTa v3 and ModernBERT achieved the most significant improvements, whereas more capable domain-specific pre-trained models like CT-BERT and BERTweet saw modest but consistent gains.

\subsubsection*{Cross Validation Results}

\begin{table}[!h]
\centering
\caption{10-fold cross validation results comparison}
\small
\begin{tabular}{lcccccc}
\toprule
\multirow{2}{*}{\textbf{Dataset}} & \multicolumn{2}{c}{\textbf{Baseline}} & \multicolumn{2}{c}{\textbf{Inceptive}} \\
\cmidrule(lr){2-3} \cmidrule(lr){4-5}
& \textbf{Mean} & \textbf{Std Dev} & \textbf{Mean} & \textbf{Std Dev} \\
\midrule
Emotion & 80.80 & 1.27 & 81.38 & 1.19 \\
Irony & 77.49 & 1.20 & 78.10 & 1.27 \\
OHSUMED & 65.06 & 1.35 & 72.57 & 0.62 \\
CAVES & 71.88 & 0.94 & 72.86 & 0.88 \\
\bottomrule
\end{tabular}
\label{table:cv}
\end{table}

\added{We conducted 10-fold cross-validation for both the baseline and inceptive models across all datasets except the large-scale Bangla dataset (resource constraints). For OHSUMED, we used the training set; for the other datasets, we combined the training and validation sets. The mean and standard deviation of the evaluation scores are reported in Table~\ref{table:cv}. Across all datasets, the inceptive models consistently achieved higher mean accuracy or F1-scores compared to the baselines. Additionally, in all but one case (irony detection), the inceptive models had a lower variance, indicating more stable performance. These results highlight the robustness and generalizability of our proposed architecture.}

\subsection{Performance vs Complexity Trade-off}\label{subsec:tradeoff}

\begin{figure}[H]
    \centering
    \includegraphics[width=.5\textwidth]{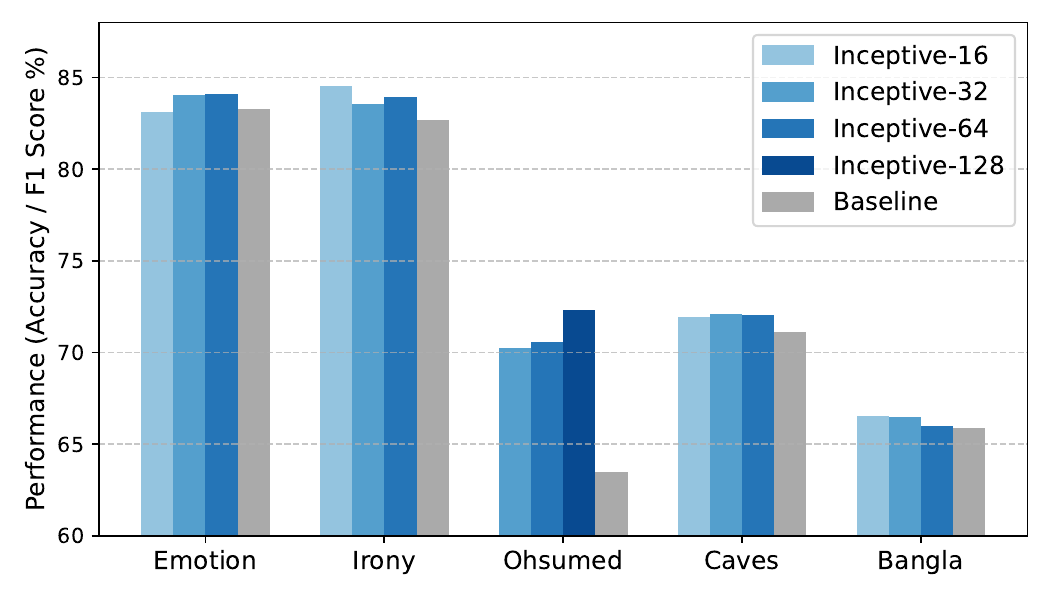} 
    \caption{Performance comparison of all tested inceptive configurations and baseline models}
    \label{fig:channels-comp}
\end{figure}

A key hyperparameter of our inceptive models is the number of output channels in convolution branches, which we tuned to determine the ideal inception module configuration in each dataset. To account for this added architectural complexity, we have compared the performance of all inception configurations against the baseline models. The results presented in Fig. \ref{fig:channels-comp} show that even the lowest performing configuration outperforms the baseline in all but one dataset, and the average performance is always higher. This suggests that extensive tuning is not strictly necessary — any selected configuration is likely to yield gain over baseline. This comparison is a post-hoc analysis performed on the test set -- these results were not used for the best configuration selection.

\noindent \textbf{A heuristic.} For shorter text sequences (emotion recognition, irony, CAVES) 16 or 32 channels perform best, while 128 channels performed best in OHSUMED only – where the text sequences are considerably larger. As a general heuristic, we recommend starting with 16 channels and scaling up based on input length or label complexity. It also aligns with the intuition that longer texts benefit from richer multi-scale features.

\subsection{Statistical Significance Testing}\label{subsec:significance}

\begin{table}[h]
    \centering
    \renewcommand{\arraystretch}{1.2}
    \caption{Wilcoxon Signed-Rank Test Results. BT: BERTweet, BB: BioBERT, RoB: RoBERTa, MB: ModernBERT, DB: DeBERTa v3, \textbf{\textit{i}}: inceptive model.}
    \small
    \begin{tabular}{p{1.4cm} p{2.2cm} c c}
        \toprule
        \textbf{Dataset} & \textbf{Models} & \textbf{Gain} & \textbf{p-value} \\
        \midrule
        \multirow{2}{*}{Emotion} 
            & BT, \textbf{\textit{i}}BT-64 & \textbf{+0.98\%} & 0.00195 \\
            & MB, \textbf{\textit{i}}MB-16 & \textbf{+3.42\%} & 0.00195 \\
        \midrule
        \multirow{3}{*}{Irony} 
            & BT, \textbf{\textit{i}}BT-16 & \textbf{+2.20\%} & 0.00585 \\
            & MB, \textbf{\textit{i}}MB-16 & \textbf{+4.33\%} & 0.00195 \\
            & DB, \textbf{\textit{i}}DB-16 & \textbf{+6.14\%} & 0.00195 \\
        \midrule
        \multirow{3}{*}{OHSUMED} 
            & BB, \textbf{\textit{i}}BB-128 & \textbf{+13.92\%} & 0.00195 \\
            & MB, \textbf{\textit{i}}MB-128 & \textbf{+12.98\%} & 0.00195 \\
            & DB, \textbf{\textit{i}}DB-128 & \textbf{+9.93\%} & 0.00195 \\
        \midrule
        \multirow{3}{*}{CAVES} 
            & RoB, \textbf{\textit{i}}RoB-32 & \textbf{+1.41\%} & 0.00195 \\
            & MB, \textbf{\textit{i}}MB-32 & \textbf{+5.24\%} & 0.00195 \\
            & DB, \textbf{\textit{i}}DB-32 & \textbf{+3.78\%} & 0.00195 \\
        \midrule
        \multirow{1}{*}{Bangla} 
            & XLM, \textbf{\textit{i}}XLM-16 & \textbf{+0.94\%} & 0.00195 \\
        \bottomrule
    \end{tabular}
    \label{tab:significance}
\end{table}

For statistical significance testing, we performed the Wilcoxon signed-rank test, which is a non-parametric test and suitable for paired comparison on the same test set. Each model was run 10 times, and the average accuracy or F1-score was recorded for statistical analysis. As shown in Table \ref{tab:significance}, the p-value in each test is below the 0.05 significance threshold. Therefore, we conclude that the gain achieved are statistically significant.

\begin{figure*}[!h]
    \centering
    \begin{subfigure}[b]{0.49\textwidth}
        \centering
        \includegraphics[width=\textwidth]{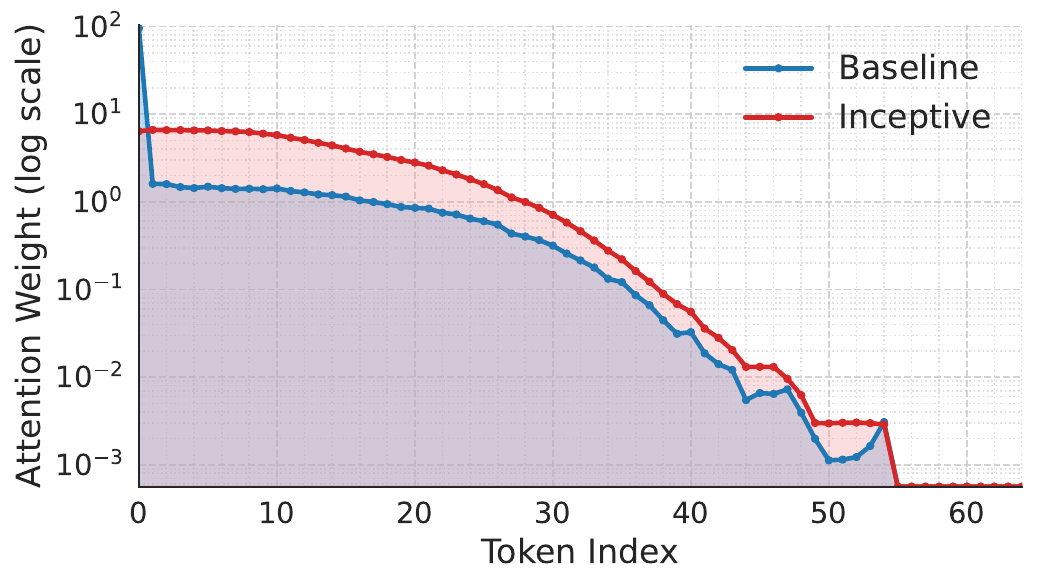}
        \caption{Baseline vs Inceptive BERTweet (Irony detection)}
        \label{fig:subfig-BT-irony}
    \end{subfigure}
    \hfill
    \begin{subfigure}[b]{0.49\textwidth}
        \centering
        \includegraphics[width=\textwidth]{visuals/attention_maps/irony_DeBERTA_baseline_vs_inceptive.pdf}
        \caption{Baseline vs Inceptive DeBERTa v3 (Irony detection)}
        \label{fig:subfig-DB-irony}
    \end{subfigure}

    \vspace{0.5em}
    \begin{subfigure}[b]{0.49\textwidth}
        \centering
        \includegraphics[width=\textwidth]{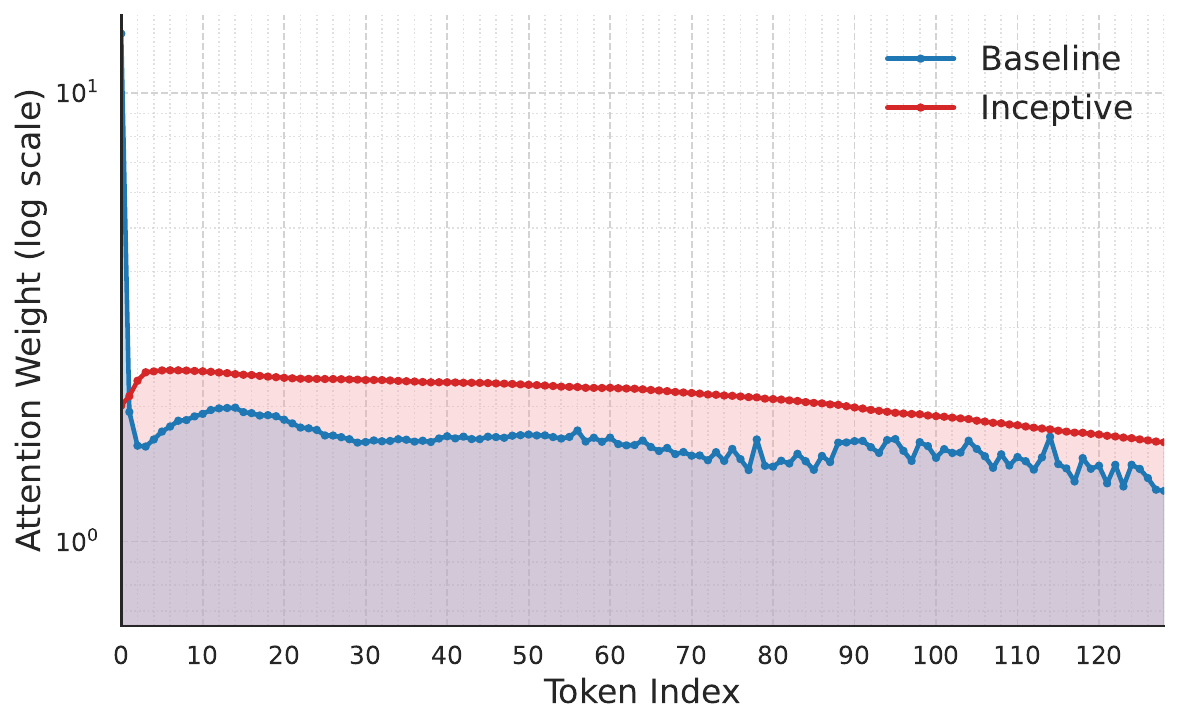}
        \caption{Baseline vs Inceptive BioBERT (OHSUMED)}
        \label{fig:subfig-BioBERT-ohsumed}
    \end{subfigure}
    \hfill
    \begin{subfigure}[b]{0.49\textwidth}
        \centering
        \includegraphics[width=\textwidth]{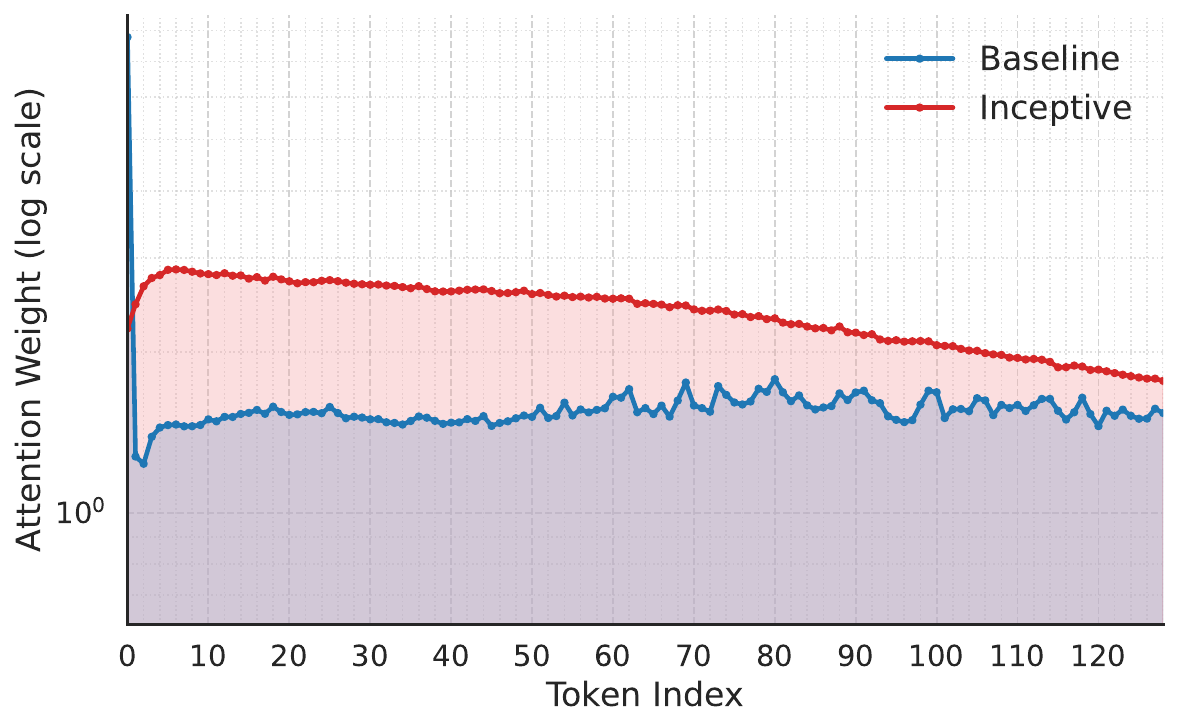}
        \caption{Baseline vs Inceptive ModernBERT (OHSUMED)}
        \label{fig:subfig-ModernBERT-ohsumed}
    \end{subfigure}

    \caption{Attention received by each token in baseline and inceptive models. Baseline attention maps are extracted from the self-attention weights of the transformer’s final layer. Inceptive attention maps come from the custom self-attention layer applied after fusion of transformers’ hidden states with multi-scale inception features. In both cases they represent token-to-token interaction weights — i.e., how much one token attends to or is influenced by others when forming the final sequence representation. Attention is visibly skewed towards the initial [CLS] token in baseline models, whereas inceptive models balance token importance dynamically based on task relevance.}
    \label{fig:attention-map-comparison}
\end{figure*}

\subsection{Performance Interpretation}\label{subsec:analysis}

The attention maps for both baseline and inceptive transformers are presented in Fig. \ref{fig:attention-map-comparison}. Attention weights in the baselines are heavily skewed toward the initial \textbf{\texttt{[CLS]}} token, while the rest of the tokens receive comparatively negligible attention. In contrast, the inceptive attention maps highlight a more balanced distribution of attention weights across the sequence. Tokens that were overlooked by transformer-based models, particularly those in the middle of the sequence, now receive more attention, reflecting their contextual importance. This is a direct result of our architectural improvements, which not only enhance contextual representations, but also enable the model to weight tokens dynamically based on their contribution to the task. 

\noindent \textbf{Disease identification.} The OHSUMED dataset involves long, complex sequences of medical abstracts, where relevant information is dispersed throughout the text. Mentions of symptoms, treatments, or diagnoses appear in different parts of the text, each contributing to the prediction of a specific disease label. As such, it is difficult for the [CLS] token to capture all these into a single token. In contrast, multi-scale convolutional branches of our inception module extract local patterns at varying granularities across the entire sequence, allowing the model to capture dependencies between neighboring tokens that may correspond to biomedical expressions, compound noun phrases, or domain-specific terminology spread throughout the abstract. As a result, our inceptive models achieve a significant \textbf{13\%} improvement over baseline in this dataset.

\noindent \textbf{Irony and emotion detection.} Irony is usually conveyed through specific linguistic patterns or subtle contextual cues like exaggerated praise or contradiction between words and context. These markers hide in nuanced phrasing rather than being explicitly stated in words -- which risks being diluted or averaged out when compressed into a single vector. In comparison, keywords like "happy" or "angry" strongly signify a particular emotion and are more likely to surface in [CLS] even with suboptimal pooling. This is why the [CLS] token proved to be “more capable” in emotion recognition (where the best-performing baseline was 0.98\% worse than inceptive model) compared to irony detection (+2.20\% gain using inception).

\subsection{Ablation Study}\label{subsec:ablation}


\begin{table*}[!h]
    \centering
    \renewcommand{\arraystretch}{1.2}
    \setlength{\tabcolsep}{8pt}
    \caption{Ablation study results for inception module, each convolution branch, self-attention, and dense block}
    \small
    \begin{tabular}{p{3.5cm} c c c c c c c c}
        \toprule
        \textbf{Model} & \textbf{Full} & \multicolumn{4}{c}{\textbf{Single convolution branch}} & \textbf{No conv} & \textbf{No attn} & \textbf{No dense} \\
        \cmidrule(lr){3-6}
         & & $k=2$ & $k=3$ & $k=5$ & $k=7$ &  &  &  \\
        \midrule
        \textbf{\textit{i}}BT (Emotion) & 84.11 & 83.82 & 83.63 & 83.06 & 83.79 & 83.27 & 83.63 & 83.51 \\
        \textbf{\textit{i}}BT (Irony) & 84.51 & 83.80 & 83.42 & 83.24 & 83.85 & 82.61 & 82.61 & 82.48 \\
        \textbf{\textit{i}}BB (OHSUMED) & 72.34 & 70.09 & 69.93 & 71.11 & 69.73 & 67.27 & 71.54 & 69.00 \\
        \textbf{\textit{i}}RoB (CAVES) & 72.11 & 71.98 & 71.68 & 71.97 & 71.49 & 71.50 & 71.31 & 71.38 \\
        \bottomrule
    \end{tabular}
    \label{tab:ablation}
\end{table*}

We perform an ablation study to understand the contribution of each main component of our framework. The results in Table \ref{tab:ablation} show that multi-scale convolution always performs better than a single convolution branch, which justifies our use of the inception module. Moreover, there is no single ‘best’ kernel size –- different kernel sizes capture complementary linguistic patterns, and their combination is what enables the model to adapt effectively across diverse domains and tasks. The ablation study also shows that both self-attention and dense block add value, as removing either worsens performance. Finally, if we remove convolution entirely, the PLM + self-attention + dense block combination performs only marginally better than the baseline PLM, proving that multi-scale feature extraction is essential.

\section{Conclusion}\label{sec:conclusion}

In this paper we presented \textit{Inceptive Transformer}, a general convolution-based framework that enhances the performance of both general-purpose transformer models like RoBERTa and domain-specific pre-trained language models such as BERTweet, BioBERT, and CT-BERT. Our experiments show statistically significant performance gains ranging from 1\% to 14\%. Moreover, our approach consistently delivers strong results across diverse domains and languages while maintaining computational efficiency. In future work, we plan to adapt our model to other tasks (e.g., NER, Q/A) and architectures (e.g., encoder-decoder models).

\section{Limitations}\label{sec:limitations}

\textbf{Dependency on output channels.} A limitation of our architecture is that it requires tuning the number of output channels in the inception module to achieve optimal performance in different datasets. For example, while an inception module with 128 output channels works best on BioBERT, 16 (for irony detection) and 32 or 64 (for emotion recognition) output channels are more suitable for BERTweet. However, we empirically found that even the lowest performing inception configuration outperformed the baseline in all but one case. 

\noindent \textbf{Focus on encoder models}. We applied our inceptive framework exclusively to bidirectional encoder-only transformer models; encoder-decoder models (e.g., T5 or BART) were not explored. Applying the inception module in such generative or sequence-to-sequence settings may require architectural adaptations.

\noindent \textbf{Impact on long-range dependencies not tested.} By design, inceptive models are most suitable for tasks that depend on local features as well as global features. We have not explored how incorporating inductive biases would impact classification on domains dominated by long-range dependencies, such as code-classification with CodeBERT style models. This can be an exciting future work. 

\section*{Acknowledgment}

We thank the anonymous reviewers and meta-reviewer for their feedback that has greatly helped in improving the quality of the paper. While writing the paper, we used AI assistance for polishing a few sentences and for some minor debugging of the code. The authors remain fully responsible for both the manuscript and the code. M Saifur Rahman is partially supported by Basic Research Grant from BUET, and CodeCrafters-Investortools Research Grant.

\bibliography{latex/acl_latex}

\appendix

\section{Baseline Architecture}

The architecture for the baseline models follows the standard fine-tuning approach for transformer-based classification tasks: the pre-trained transformer layer is followed by a dropout layer (p = 0.3) and a linear classification head. The hidden state of the \textbf{\texttt{[CLS]}} token from the final transformer layer is used as the sequence representation, which is passed through dropout to prevent overfitting and then mapped to the output classes via a fully connected layer.

\section{Dataset Class Distributions}\label{appendix:datasets}

\begin{table}[H]
\renewcommand{\arraystretch}{1.2}
\centering
\caption{Dataset statistics. $C:$ number of classes or labels; 
$\overline{C}:$ average number of labels per instance (for multi-label); 
and $\overline{L}:$ average token length of each text.}
\begin{tabular}{lrrrr}
\toprule
\textbf{Dataset} & \textbf{\#Texts} & $\mathbf{C}$ & $\boldsymbol{\overline{C}}$ & $\boldsymbol{\overline{L}}$ \\
\midrule
Emotion & 5,052  & 4  & --   & 24.35 \\
Irony   & 4,601  & 2  & --   & 21.54 \\
Bangla   & 80,098 & 6 & -- & 18.6 \\
OHSUMED & 13,929 & 23 & 1.66 & 289.51 \\
CAVES   & 9,921 & 11 & 1.16 & 58.35 \\
\bottomrule
\end{tabular}
\label{tab:data_stats}
\end{table}

\begin{figure}[H]
    \centering
    \includegraphics[width=.4\textwidth]{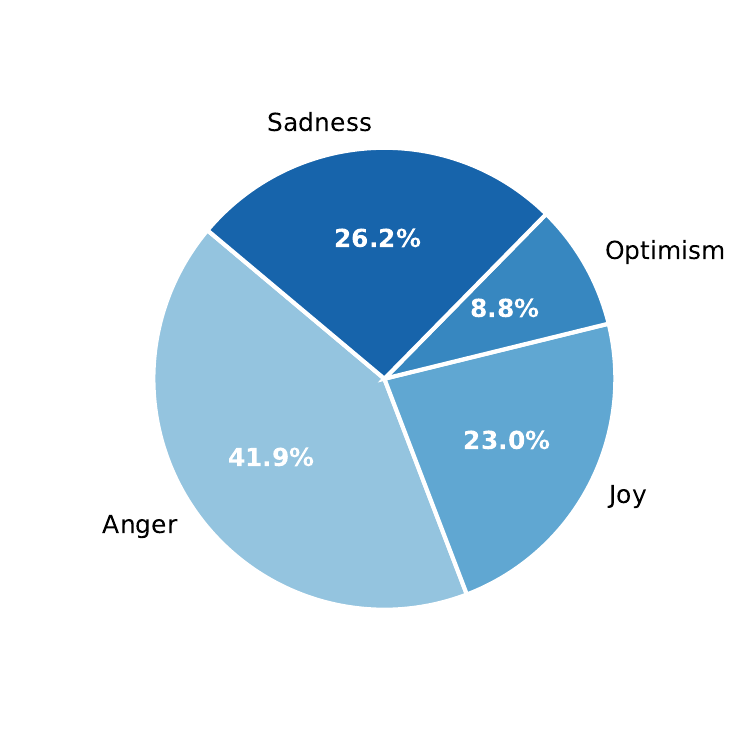} 
    \caption{Class distribution in Emotion Recognition}
    \label{fig:emotion-class}
\end{figure}

\begin{figure}[H]
    \centering
    \includegraphics[width=.4\textwidth]{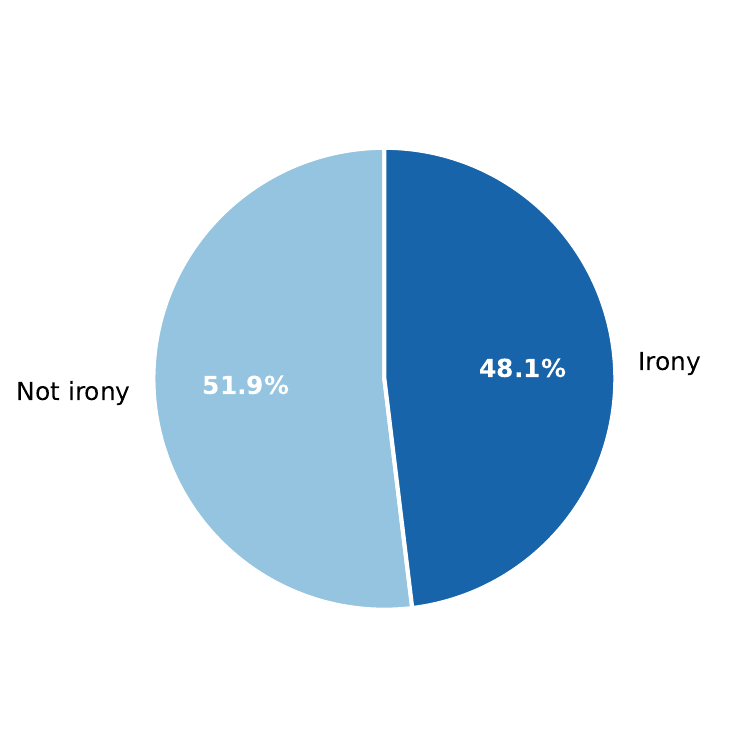} 
    \caption{Class distribution in Irony Detection}
    \label{fig:irony-class}
\end{figure}

\begin{figure}[H]
    \centering
    \includegraphics[width=.4\textwidth]{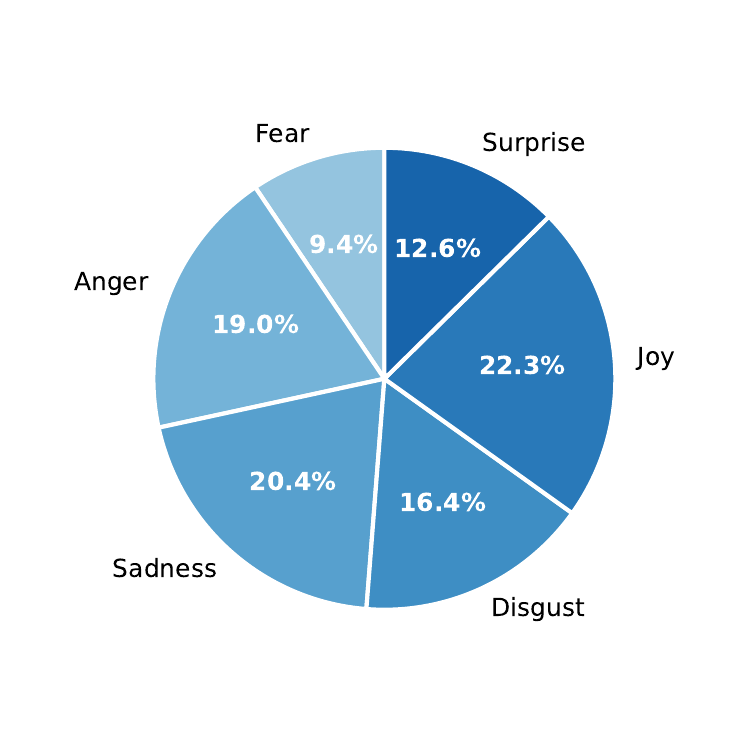} 
    \caption{Class distribution in Bangla emotion detection}
    \label{fig:bangla-class}
\end{figure}

\begin{figure}[H]
    \centering
    \includegraphics[width=.45\textwidth]{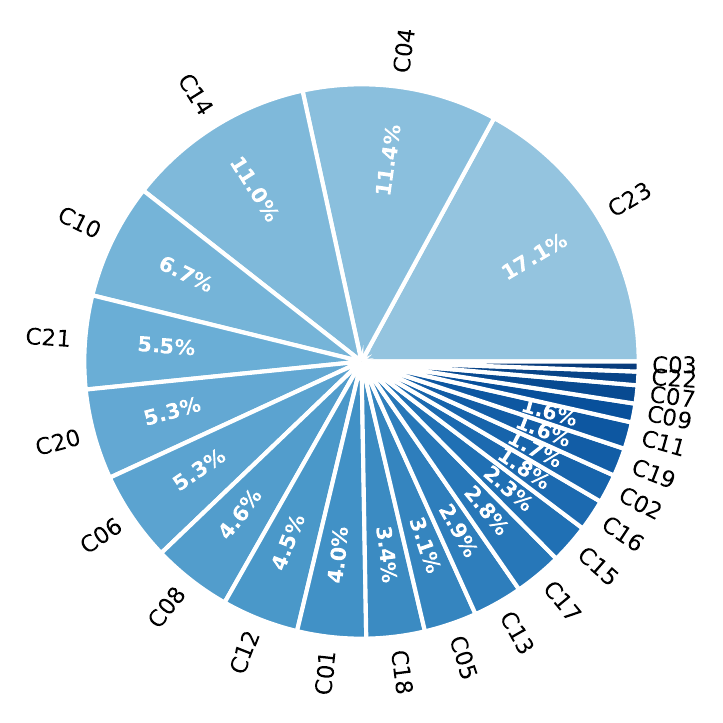} 
    \caption{Class distribution in OHSUMED}
    \label{fig:ohsumed-class}
\end{figure}

\begin{figure}[H]
    \centering
    \includegraphics[width=.45\textwidth]{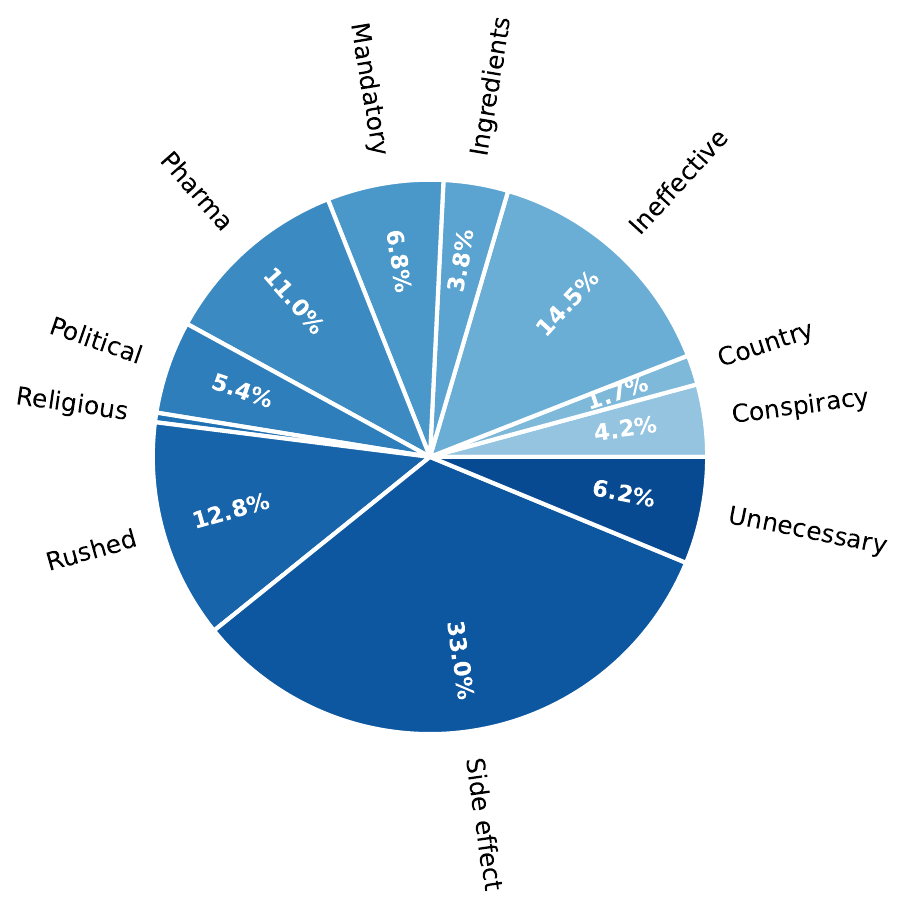} 
    \caption{Class distribution in CAVES}
    \label{fig:caves-class}
\end{figure}

\section{Word clouds}

\begin{figure}[H]
    \centering
    \includegraphics[width=.4\textwidth]{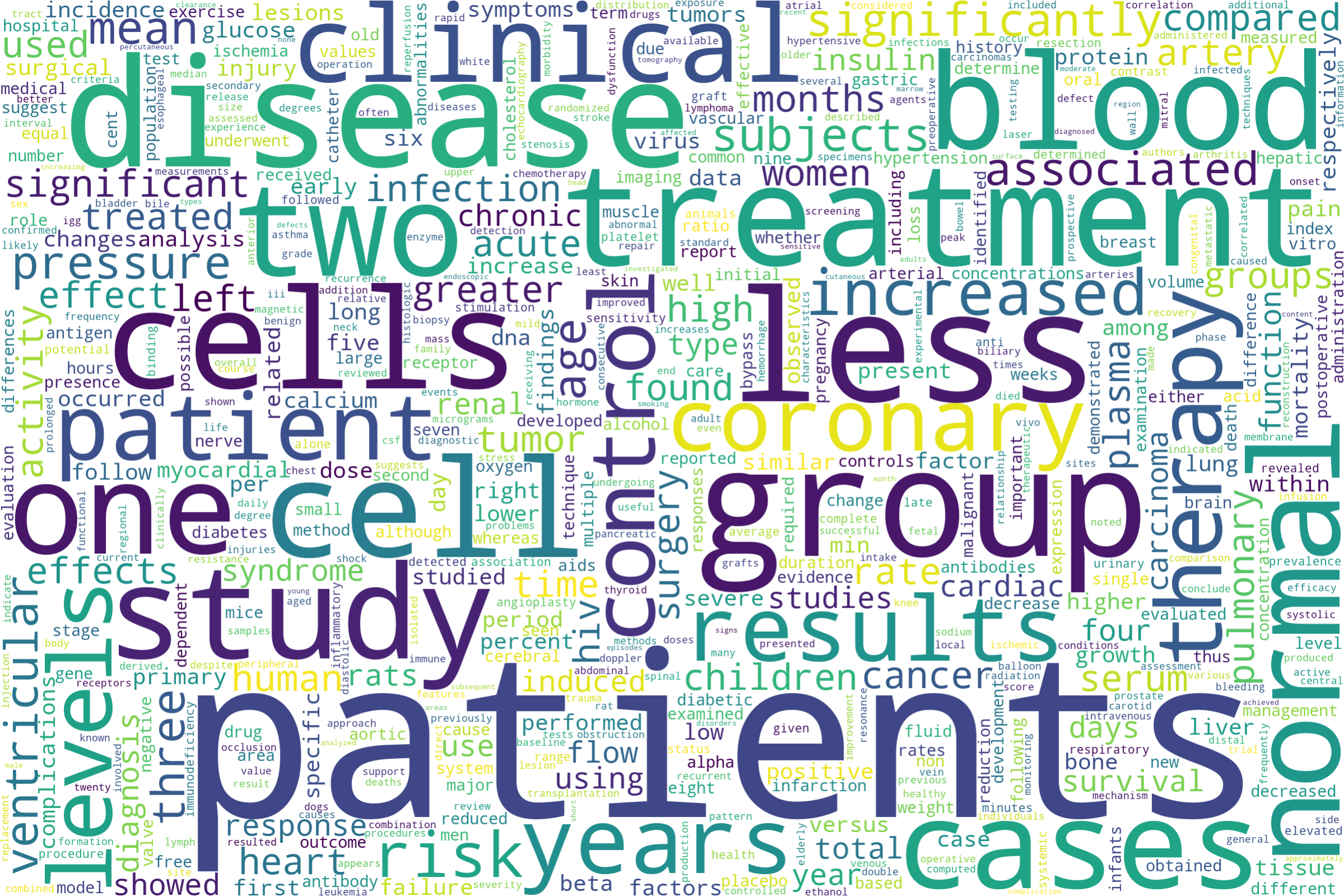} 
    \caption{Word clouds of OHSUMED dataset}
\end{figure}

\begin{figure}[H]
    \centering
    \includegraphics[width=.4\textwidth]{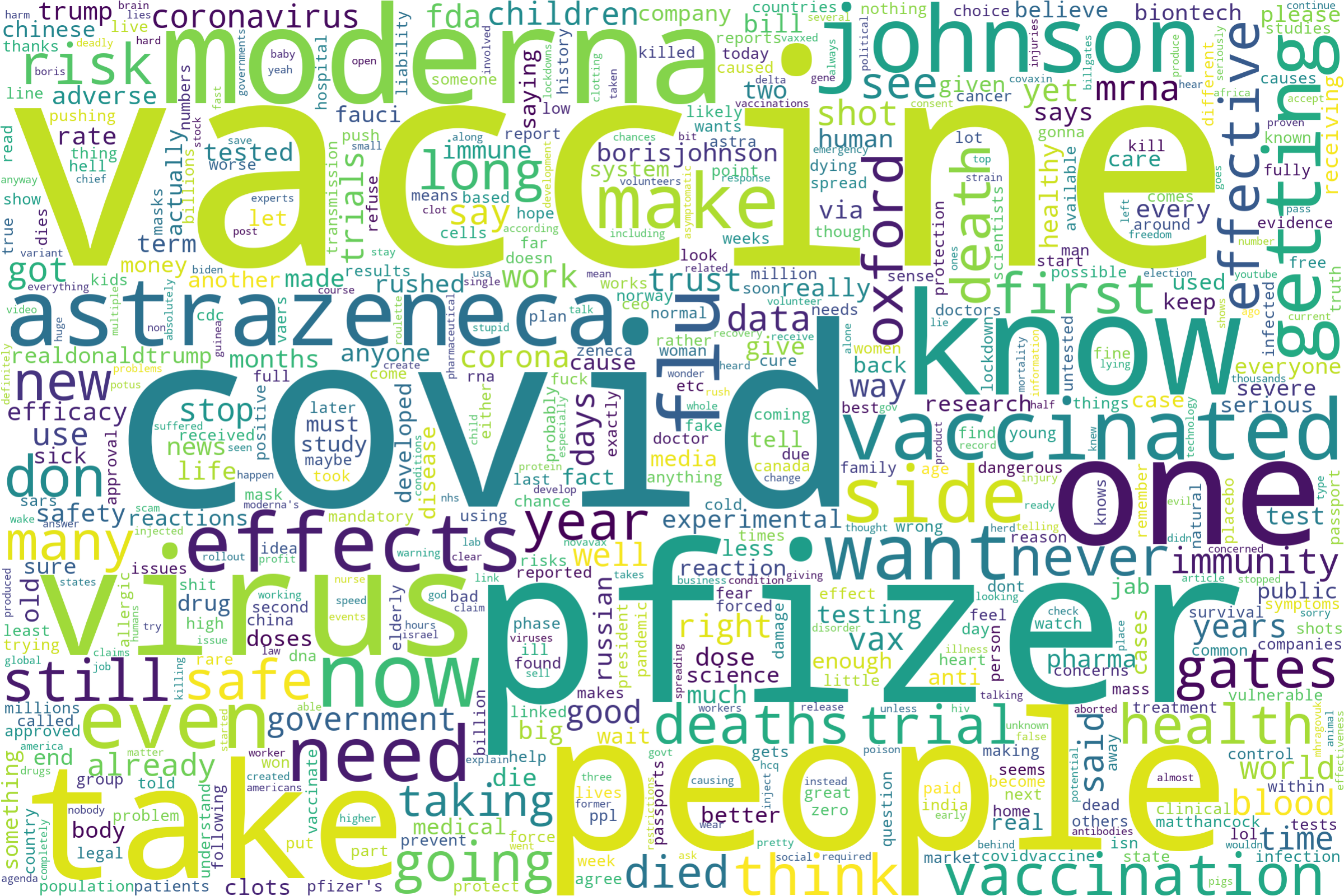} 
    \caption{Word clouds of CAVES dataset}
\end{figure}

\begin{figure}[H]
    \centering
    \includegraphics[width=.4\textwidth]{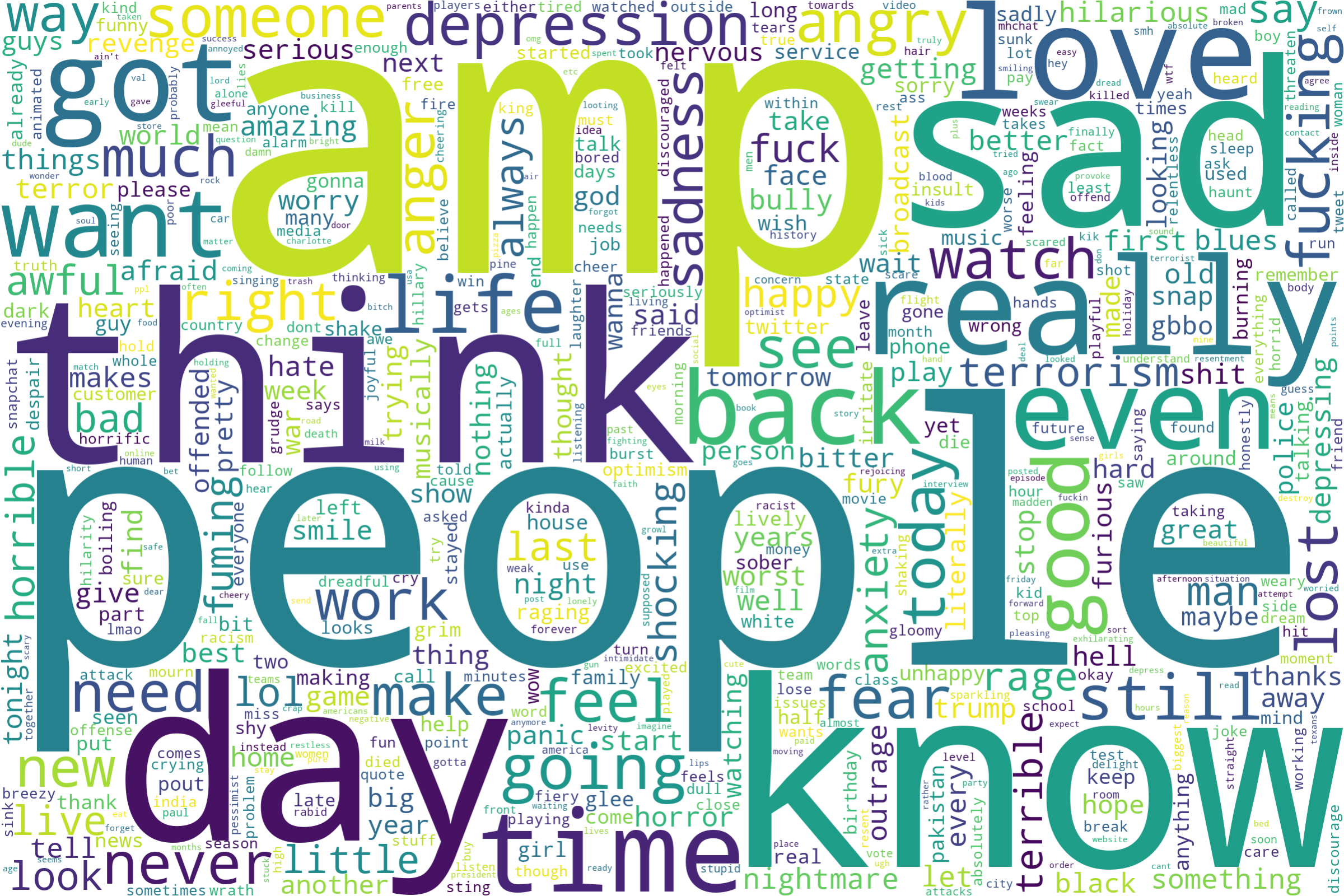} 
    \caption{Word clouds of emotion recognition dataset}
\end{figure}

\begin{figure}[H]
    \centering
    \includegraphics[width=.4\textwidth]{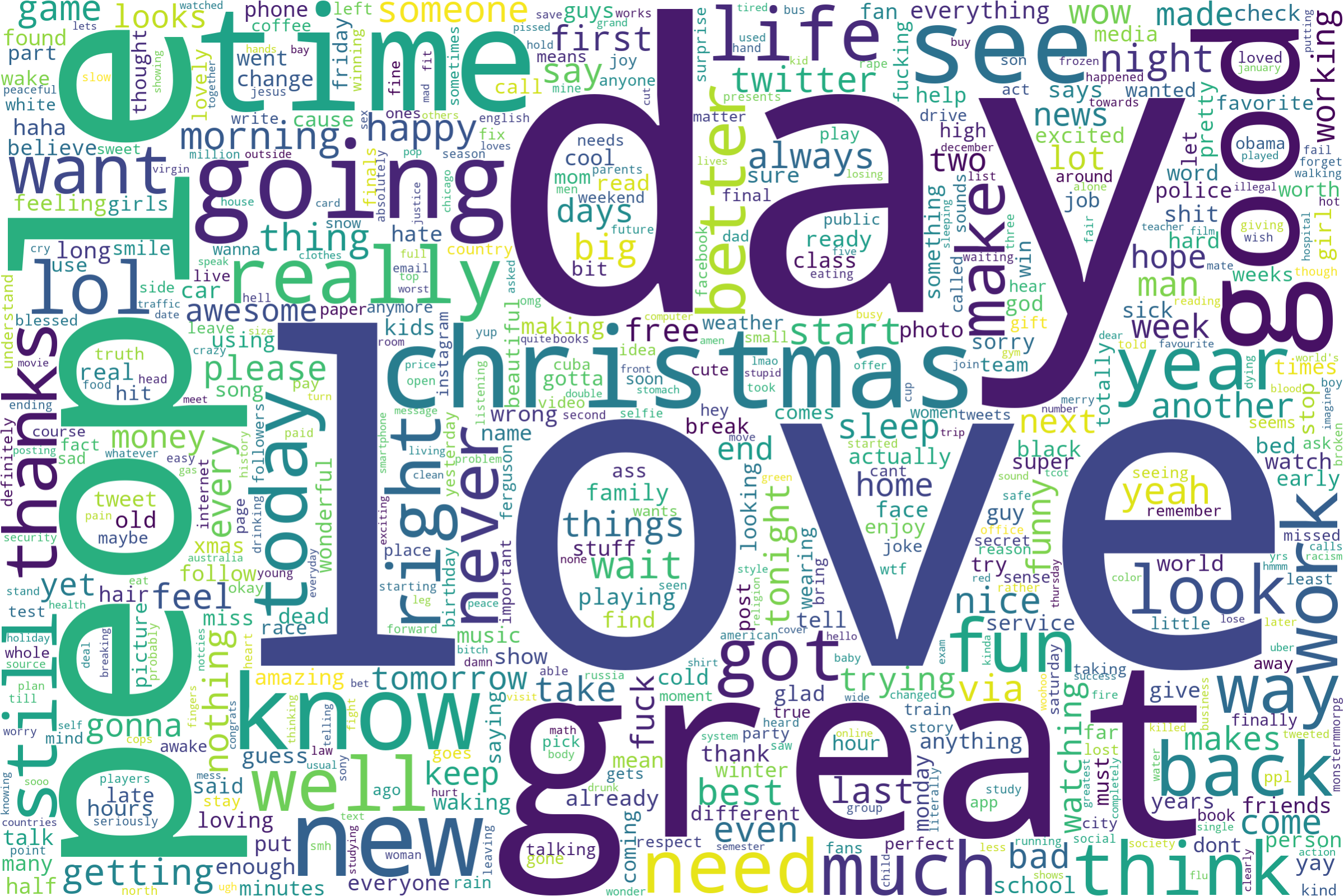} 
    \caption{Word clouds of irony detection dataset}
\end{figure}

\section{Full Performance Comparison}
\label{sec:appendix:full-perf}

\begin{figure}[H]
    \centering
    \includegraphics[width=.5\textwidth]{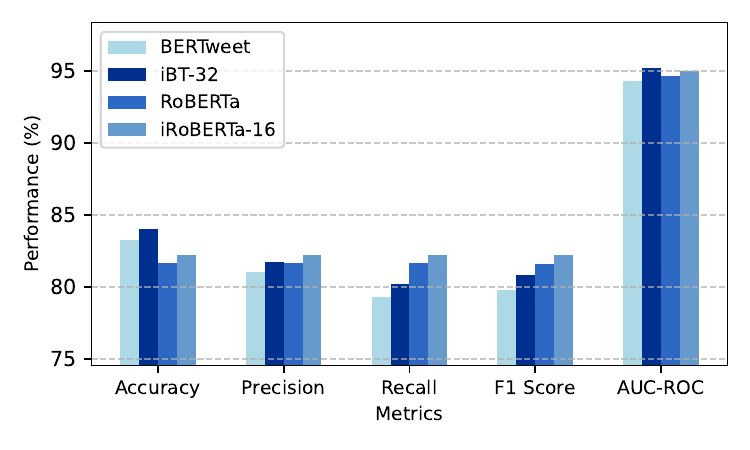} 
    \caption{Performance comparison in Emotion Recognition}
\end{figure}

\begin{figure}[H]
    \centering
    \includegraphics[width=.5\textwidth]{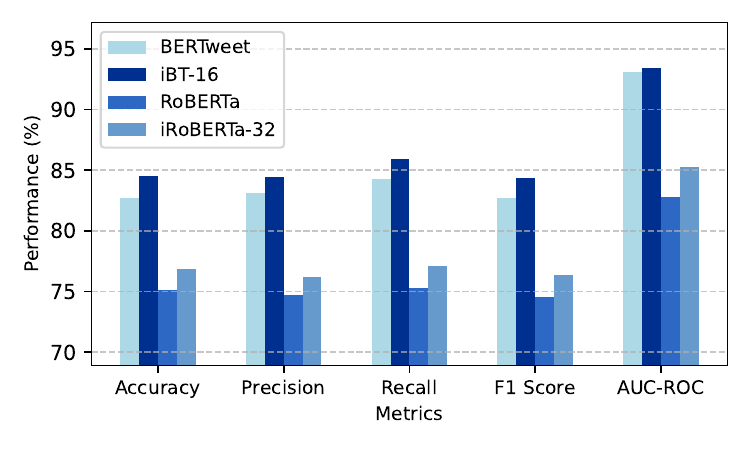} 
    \caption{Performance comparison in Irony Detection}
\end{figure}

\begin{figure}[H]
    \centering
    \includegraphics[width=.5\textwidth]{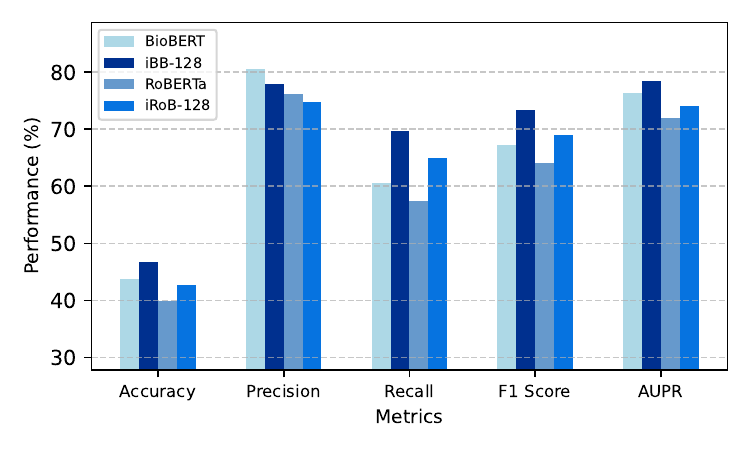} 
    \caption{Performance comparison in OHSUMED}
\end{figure}

\begin{figure}[H]
    \centering
    \includegraphics[width=.5\textwidth]{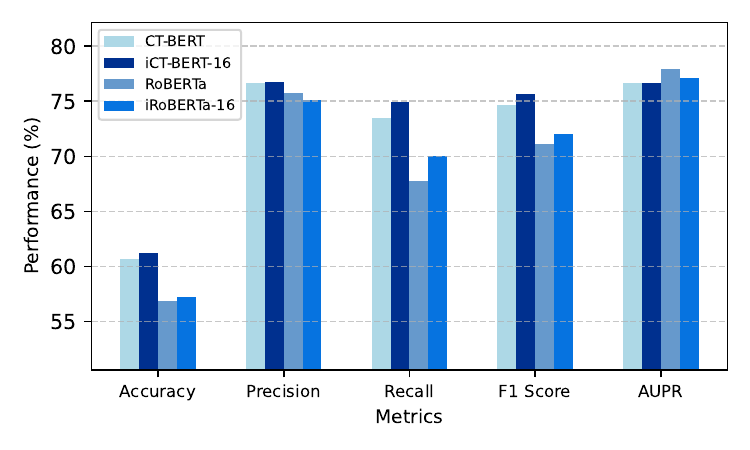} 
    \caption{Performance comparison in CAVES}
\end{figure}

\section{Comparison across All Runs}\label{sec:appendix:all-runs}

\begin{figure}[H]
    \centering
    \includegraphics[width=.45\textwidth]{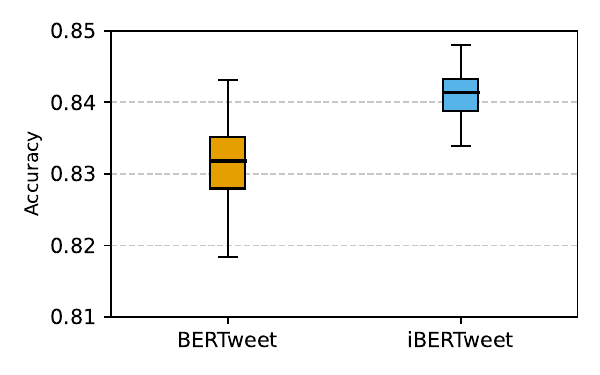} 
    \caption{Accuracy distribution across 10 runs in Emotion Recognition}
    \label{fig:emotion-all-runs}
\end{figure}

\begin{figure}[H]
    \centering
    \includegraphics[width=.45\textwidth]{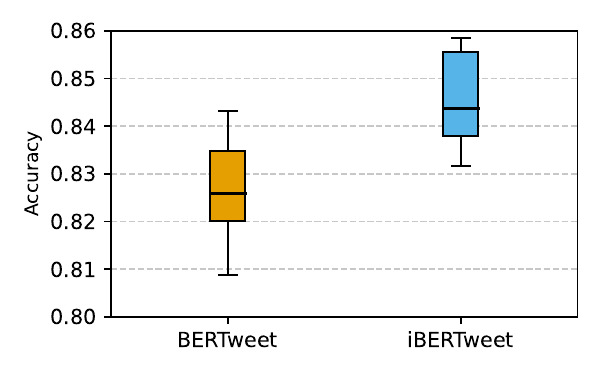} 
    \caption{Accuracy distribution across 10 runs in Irony Detection}
    \label{fig:irony-all-runs}
\end{figure}

\begin{figure}[H]
    \centering
    \includegraphics[width=.45\textwidth]{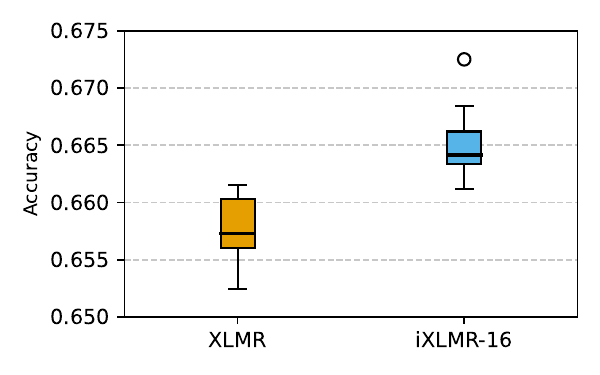} 
    \caption{Accuracy distribution across 10 runs in Bangla emotion detection}
    \label{fig:bangla-all-runs}
\end{figure}

\begin{figure}[H]
    \centering
    \includegraphics[width=.45\textwidth]{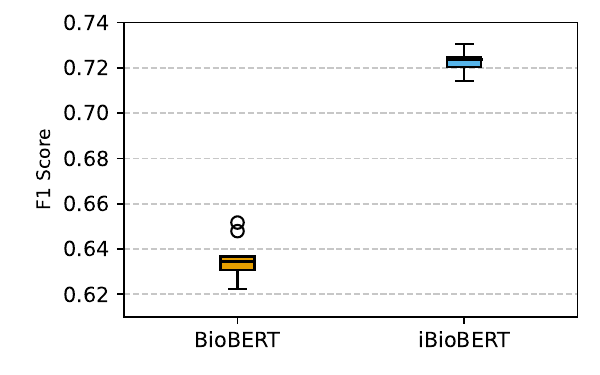} 
    \caption{F1-score distribution across 10 runs in OHSUMED}
    \label{fig:ohsumed-all-runs}
\end{figure}

\begin{figure}[H]
    \centering
    \includegraphics[width=.45\textwidth]{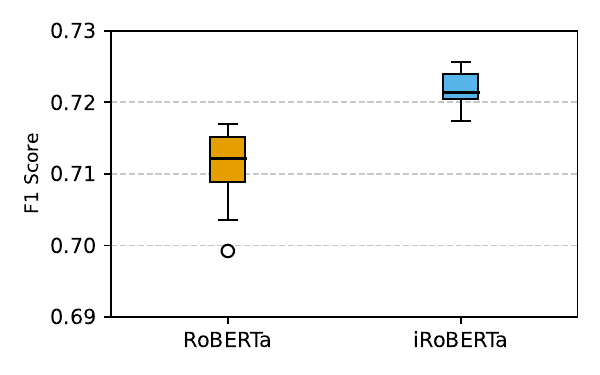} 
    \caption{F1-score distribution across 10 runs in CAVES}
    \label{fig:caves-all-runs}
\end{figure}

Fig. \ref{fig:emotion-all-runs}, \ref{fig:irony-all-runs}, \ref{fig:bangla-all-runs}, \ref{fig:ohsumed-all-runs}, and \ref{fig:caves-all-runs} show the comparison of baseline pretrained models (BERTweet, XLMR, BioBERT, RoBERTa) against the inception models across all 10 runs.

\section{Attention Maps}\label{sec:appendix:attention_map}

\begin{figure}[H]
    \centering
    \includegraphics[width=.5\textwidth]{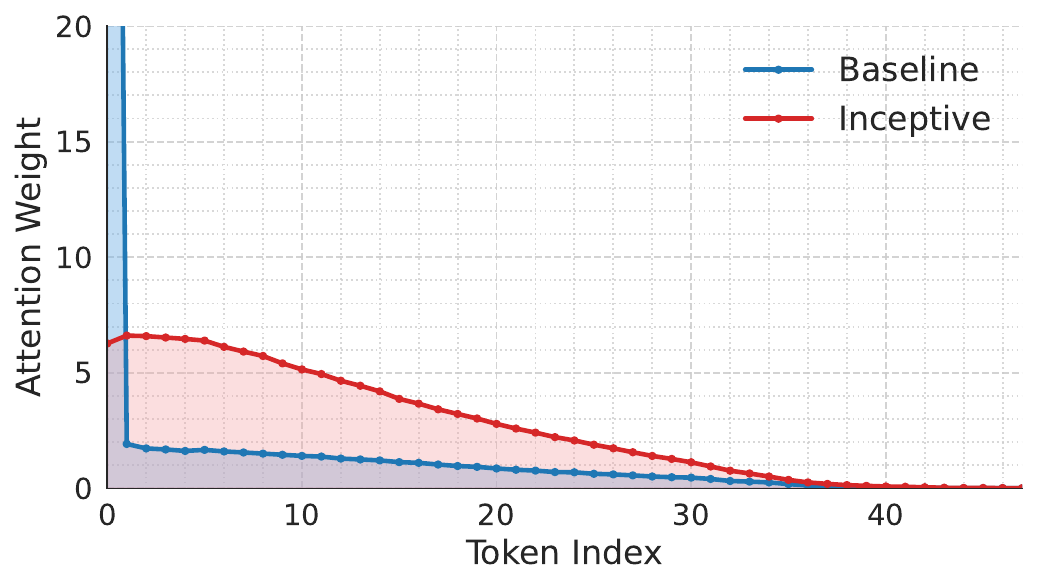} 
    \caption{Baseline vs Inceptive BERTweet attention map (emotion)}
    \label{fig:emotion_BT}
\end{figure}

\begin{figure}[H]
    \centering
    \includegraphics[width=.5\textwidth]{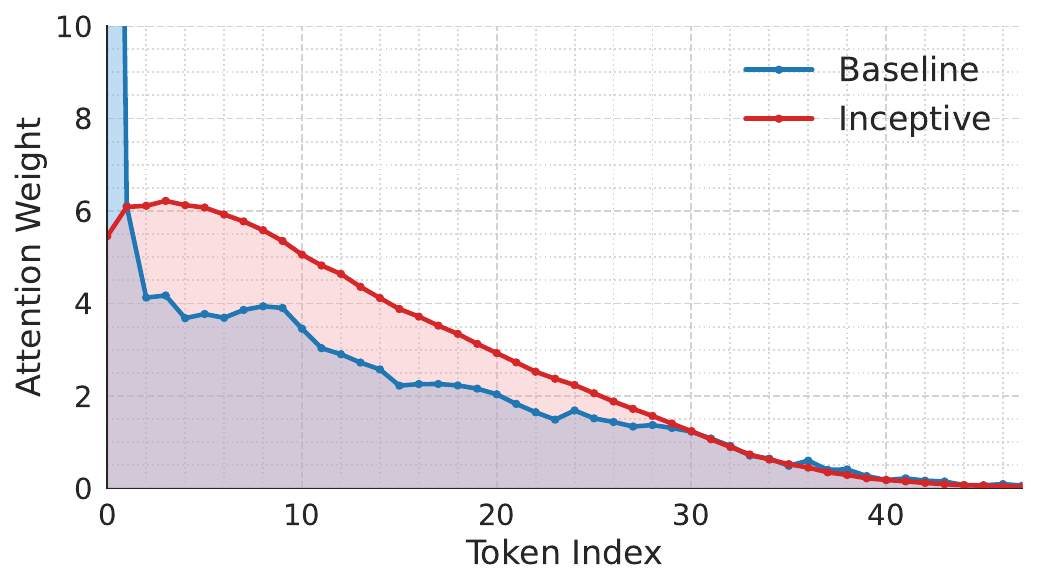} 
    \caption{Baseline vs Inceptive RoBERTa attention map  (emotion)}
    \label{fig:emotion_RoB}
\end{figure}

\begin{figure}[H]
    \centering
    \includegraphics[width=.5\textwidth]{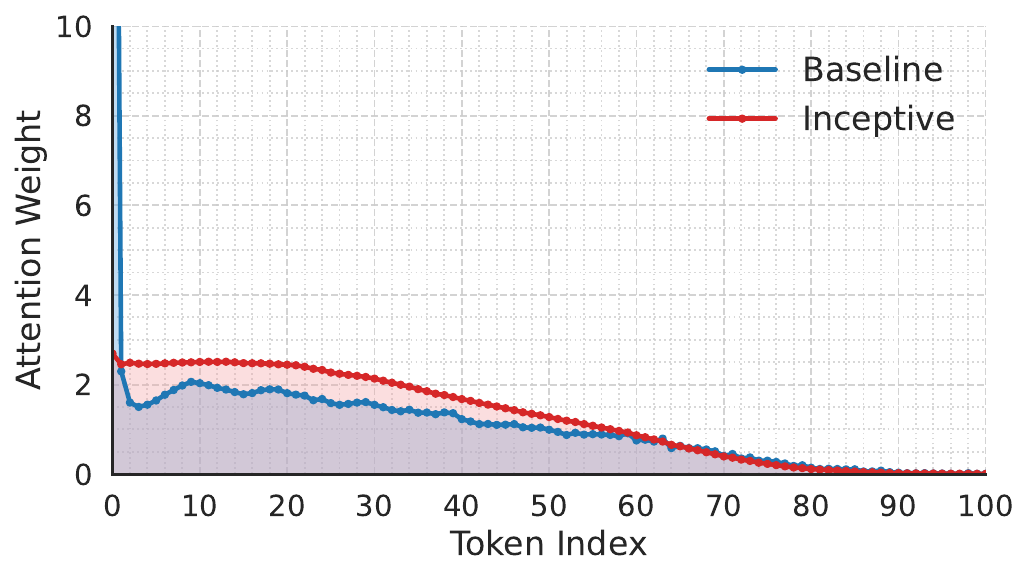} 
    \caption{Baseline vs Inceptive RoBERTa attention map  (CAVES)}
    \label{fig:caves_rob}
\end{figure}

\begin{figure}[H]
    \centering
    \includegraphics[width=.5\textwidth]{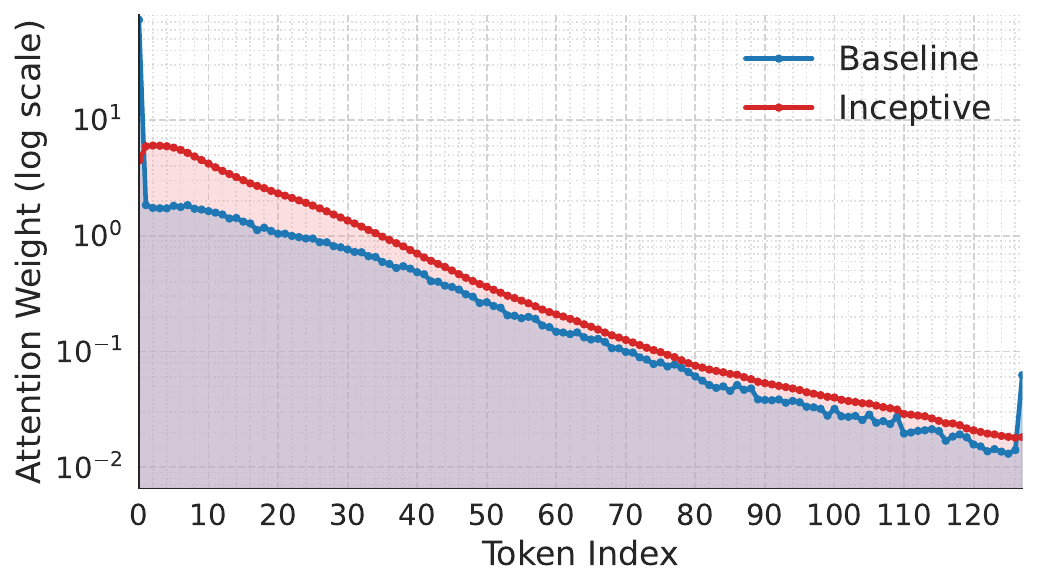} 
    \caption{Baseline vs Inceptive XLM-R attention map (Bangla)}
    \label{fig:bangla_XLMR}
\end{figure}

\section{Stacking Multiple Inception Modules}

We implemented a simple variant of our architecture that stacks two inception modules sequentially -- the output of the first inception module is fed into the second one for further multi-scale refinement. We tested it on the two datasets where we had most improvements -- irony and OHSUMED -- and found that it performs equally or slightly worse than our original architecture (but still better than the baselines). This is likely due to the additional module introducing redundant or over-smoothed representations, which may obscure the fine-grained features captured by the first inception layer. 

\begin{figure}[H]
    \centering
    \includegraphics[width=.5\textwidth]{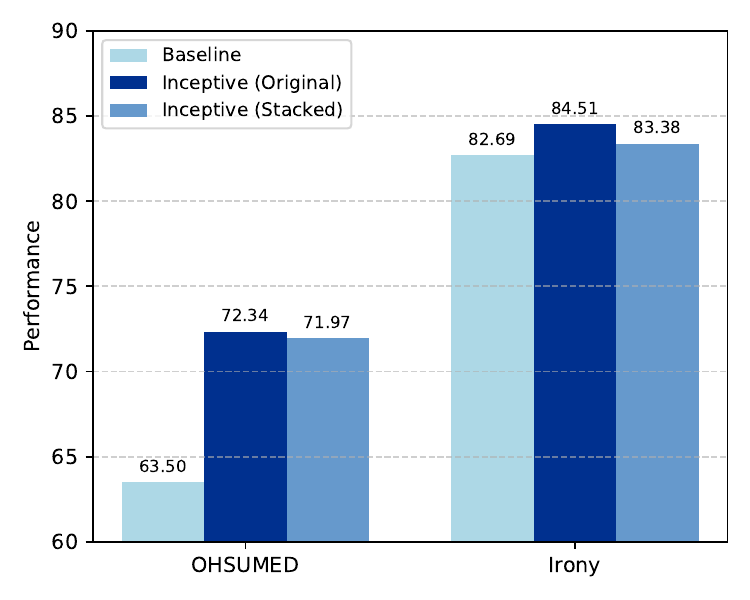} 
    \caption{Single inception vs 2-stacked inception}
    \label{fig:caves_incROB}
\end{figure}

\end{document}